\documentclass{article}
\usepackage{spconf,amsmath,epsfig}
\usepackage[linesnumbered,boxed]{algorithm2e}
\usepackage{multirow}
\usepackage{booktabs}
\usepackage{subfigure}
\usepackage{xcolor}
\usepackage[colorlinks,linkcolor=blue,citecolor=black]{hyperref}
\begin{document}\sloppy

\def\x{{\mathbf x}}
\def\L{{\cal L}}

\title{WAVELET CHANNEL ATTENTION MODULE WITH A FUSION NETWORK FOR SINGLE IMAGE DERAINING}
%
\name{Hao-Hsiang Yang$^{1}$, Chao-Han Huck Yang$^{2}$, Yu-Chiang Frank Wang$^{1,3}$}
\address{$^{1}$ ASUS Intelligent Cloud Services, Taiwan \\ $^{2}$School of Electrical and Computer Engineering, Georgia Institute of Technology, Atlanta, GA, USA \\ $^{3}$Department of Electrical Engineering, National Taiwan University, Taiwan}

\maketitle

\begin{abstract}

\end{abstract}

Single image deraining is a crucial problem because rain severely degenerates the visibility of images and affects the performance of computer vision tasks like outdoor surveillance systems and intelligent vehicles. In this paper, we propose the new convolutional neural network (CNN) called the wavelet channel attention module with a fusion network. Wavelet transform and the inverse wavelet transform are substituted for down-sampling and up-sampling so feature maps from the wavelet transform and convolutions contain different frequencies and scales. Furthermore, feature maps are integrated by channel attention. Our proposed network learns confidence maps of four sub-band images derived from the wavelet transform of the original images. Finally, the clear image can be well restored via the wavelet reconstruction and fusion of the low-frequency part and high-frequency parts. Several experimental results on synthetic and real images present that the proposed algorithm outperforms state-of-the-art methods.

\section{Introduction}
Single image deraining is a crucial problem because rain occludes the background scene, appears in different locations and decreases the performance of computer vision tasks like outdoor surveillance systems and intelligent vehicles \cite{Li_2019_CVPR}. The rain streaks can be seen as linear noises which may vary in size, direction and density. Moreover, in real cases, when rain accumulation is dense, the individual streaks cannot be observed clearly. Accumulated rain streaks reduce the visibility in a manner more similar to fog, and create a haze-like phenomenon in the background. This foggy phenomenon can be described as the haze model \cite{mccartney1976optics}, and the whole model \cite{li2019heavy} is written as
\begin{equation}
J=T\odot(I+\sum_{i}^{n}S_{i})+(1-T)\odot A
\label{eq:res}
\end{equation}
where $J$ means the observed rain streak image, $I$ means the clear image, $S_{i}$ means a rain streak layer with the same direction, $n$ means the maximum number of layers, $A$ means the global atmospheric light, $T$ means the atmospheric transmission and $\odot$ denotes element-wise multiplication. 

Several methods try to analyze visual priors to capture deterministic and statistical properties of rainy images \cite{chen2013generalized,luo2015removing,ronneberger2015u}. However, these methods tend to introduce undesirable artifacts, since their handcrafted priors from human observations do not always hold in diverse real-world images.
Instead of applying handcrafted visual priors, recently, deep-learning-based methods \cite{li2019heavy,pan2018learning,Zhang_2018_CVPR,fu2019lightweight} are proposed, and these methods usually perform more accurate than conventional priors-based methods with significant performance gains. 

We also consider neural networks for single image deraining since the deraining model is a crude approximation. CNN-based model can learn and capture more detailed features from rainy images.
Similar to other image restoration tasks like image deblurring \cite{pan2018learning}, and image dehazing \cite{yang2019wavelet}, image deraining can be modeled as an image-to-image mapping problem. 
From previous studies~\cite{Zhang_2018_CVPR, fu2019lightweight, yang2019wavelet}, low-level features (e.g., edge and frequency) are more important than high-level features like attribute, texture.
To extract these low-level features, we apply the wavelet transform and propose the wavelet channel attention module (WCAM) with a fusion network. First, our network replaces the down-sampling and up-sampling with the discrete wavelet transform (DWT) and the inverse discrete wavelet transform (IDWT). The network further captures the various frequency features and bi-orthogonal property of the DWT for signal recovery. 
Second, the channel attention \cite{hu2018squeeze} module selectively emphasizes interdependent channel maps by integrating associated features among all channel maps. Thus, we combine the DWT and the channel attention so that intermediate feature maps with different frequencies are integrated effectively. Third, as demonstrated in \cite{Zhang_2018_CVPR,Ren_2018_CVPR}, fusing various levels of features is beneficial for many computer vision tasks. The DWT is seen as the fusion of high-frequency and low-frequency images from original images. Inspired by it, our input is four sub-band images from the DWT, and the output is four confidence maps that determine the importance of different sub-band images and fuse them to reconstruct the clear image by the IDWT. In summary, Our network is encoder-decoder architecture. The proposed WCAMs replace convolutions in the encoder, and corresponding inverse wavelet channel attention modules (IWCAMs) also replace convolutions in the decoder.

This paper makes the following contributions: (i) We propose a novel end-to-end WCAM with a fusion network that captures frequency features and fuses four sub-band images for single image deraining. (ii) Several experiments show that the proposed network obtains much better performance than previous state-of-the-art deraining methods, even in the presence of large rain streaks and rain streak accumulation.

\vspace{-0.3cm}
\section{Related Work}
\subsection{Single image deraining}
Single image deraining is an ill-posed problem and seen as the denoising problem. Early methods propose hand-craft priors to estimate distributions of rain streaks and remove them. In \cite{chen2013generalized}, Chen and Hsu decompose the background and rain streak layers based on low-rank priors to restore clear images. Luo \emph{et al.} \cite{luo2015removing} use sparse coding with high discriminability so that the derained image layer and the rain layer can be accurately separated. In \cite{li2016rain}, patch-based priors are applied for both the clean background and rain layers in the form of Gaussian mixture models to remove rain streaks. 

Recently, trainable CNNs have been proposed to estimate the clean image from a rainy input directly. In \cite{pan2018learning}, a dual convolutional neural network for deraining is proposed. The proposed network consists of two parallel branches, which respectively recovers the structures and details in an end-to-end manner. Zhang \emph{et al.} \cite{Zhang_2018_CVPR} propose a density-aware multi-stream densely connected convolutional neural network that jointly estimates rain density estimation and derained image. In \cite{fu2019lightweight}, the authors introduce the Gaussian-Laplacian image pyramid decomposition technology to the neural network and propose a lightweight pyramid network for single image deraining. In \cite{li2019heavy}, the authors propose the two-stage network. Rain streaks, transmissions, and the atmospheric light are estimated in the first stage. Derained images are refined in the second stage by the conditional generative adversarial network. Different from them, our network is implemented in the wavelet space to capture detailed frequency information.
\subsection{Attention mechanisms}
Attention plays an important role in human perception and computer vision tasks. Attention mechanisms give feature maps weights so that features of the sequence of regions or locations are magnified. Generally, there are two attention mechanisms: spatial attention and channel attention \cite{woo2018cbam, yang2020characterizing}. Mnih et al. \cite{mnih2014recurrent} propose an attention model that spatially selects a sequence of regions to refine feature maps, and the network not only performs well but is also robust to noisy inputs. Hu et al. \cite{hu2018squeeze} propose the squeeze-and-excitation module and use global average-pooled features to compute channel-wise attention. Furthermore, Woo et al.\cite{woo2018cbam} combine the spatial and channel attention to propose a convolutional block attention module. Their module sequentially infers attention maps along two separate dimensions, channel and spatial, then attention maps are multiplied to the input feature map for adaptive feature refinement, which increases the accuracy of image recognition. In our work, we integrate channel attention and wavelet transform so that output feature maps contain frequency features but different magnified weights.
\vspace{-0.3cm}
\section{PROPOSED METHODS}
\subsection{Wavelet Channel Attention module}
We first describe the 2-D discrete wavelet transform (DWT) in our model. We apply Haar wavelets \cite{mallat1999wavelet} and it contains four kernels, $LL^{\top}, HL^{\top}, LH^{\top}, HH^{\top}$, where $L$ and $H$ are the low pass filter and high pass filter respectively. Both filters are
\begin{equation}
L= \frac{1}{\sqrt2}[1,1]^{\top}, H = \frac{1}{\sqrt2}[1, -1]^{\top}
\label{eq:res}
\end{equation}
The DWT for image processing means an image $I$ is convolved and then down-sampled to obtain the four sub-band images $I_{LL}, I_{HL}, I_{LH}$ and $I_{HH}$. The low-pass filter captures smooth surface and texture while the three high-pass filters extract vertical, horizontal, and diagonal edge-like
information. 
Even though the downsampling operation is employed, due to the biorthogonal property of DWT, the original image $I$ can be accurately reconstructed by the inverse discrete wavelet transform (IDWT).
Therefore, the DWT is seen as four $2\times2$ convolutional kernels whose weights are fixed and the stride equals two. Similarily, the IDWT is seen as the transpose convolution operation whose kernels are identical to DWT's.

Then, we extend convolutions by combining the DWT and the channel attention mechanism and propose the wavelet channel attention module (WCAM). A WCAM is a computational unit that is built upon a transformation $F$ mapping an
intermediate feature map $ X\in R^{C\times H\times W} $ to a feature maps $ Y\in R^{4C\times \frac{H}{2}\times \frac{W}{2}} $. Given an intermediate feature map $ X\in R^{C\times H\times W} $, the DWT decomposes $X$ into ${\rm DWT}(X)=[X_{LL}, X_{HL}, X_{LH}, X_{HH}] \in R^{4C\times\frac{H}{2}\times \frac{H}{2}}$. $3\times3$ convolutions and leaky rectified linear units (LeakyReLUs) are applied to extract various frequency features from DWT$(X)$ and denoted as $Conv({\rm DWT}(X))$, where $Conv$ is an operator combining the convolution and the LeakyReLU. Furthermore, we propose improved feature maps with different frequencies contributing different weights to restore the clear image. We use the channel attention \cite{hu2018squeeze} to control weights of various channel-wise features. The proposed module calculates the global average pooling of $X$ and uses $1\times1$ convolutions to infer a channel attention map $ M_{c}\in R^{4C\times 1\times 1}$. Therefore, the entire result of WCAM is $Conv( {\rm DWT}(X)) \odot M_{c}$. The detailed structure of the WCAM is depicted in Fig. 2(b) and formulated as follows:
\begin{equation}
\begin{split}
M_{c} &= Conv( {\rm AP}(X))\\
Y&=F(X) = Conv({\rm DWT}(X)) \odot M_{c} 
\label{eq:WCAM}
\vspace{-0.7cm}
\end{split}
\end{equation}
where AP means global average pooling.

The WCAM reduces the size of feature maps but increases the receptive field to capture multi-frequency and multi-scale features. To magnify the size of the feature map, similarly, the inverse wavelet channel attention module (IWCAM) is proposed. An IWCAM is also a computational unit which can be built upon a transformation $F^{-1}$ mapping an intermediate feature map $ X\in R^{4C\times \frac{H}{2}\times \frac{W}{2}} $ to a feature map $ Y\in R^{C\times H\times W} $. Given an intermediate feature map $ X\in R^{4C\times \frac{H}{2}\times \frac{W}{2}} $, the IDWT merges $X$ into $ {\rm IDWT}(X) \in R^{C\times H\times W}$ and $3\times3$ convolutions and LeaklyReLUs are then adopted and denoted as $Conv( {\rm IDWT}(X)) \in R^{C\times H\times W}$. Global average pooling of $X$ is calculated and $1\times1$ convolutions are used to infer a channel attention map $ M_{c}\in R^{4C\times 1\times 1}$. The entire result of IWCAM is $Conv( {\rm IDWT}(X)) \odot M_{c}$. The structure of the IWCAM is depicted in Fig. 2(c) and formulated as follows:
\begin{equation}
\begin{split}
M_{c} &= Conv( {\rm AP}(X))\\
Y&=F^{-1}(X)= Conv({\rm IDWT}(X)) \odot M_{c}
\label{eq:IWCAM}
\end{split}
\end{equation}
\vspace{-0.8cm}

\begin{figure}[!htb]
\centering

\subfigure[]{
\begin{minipage}[t]{1\linewidth}
\centering
\includegraphics[width=8.5cm]{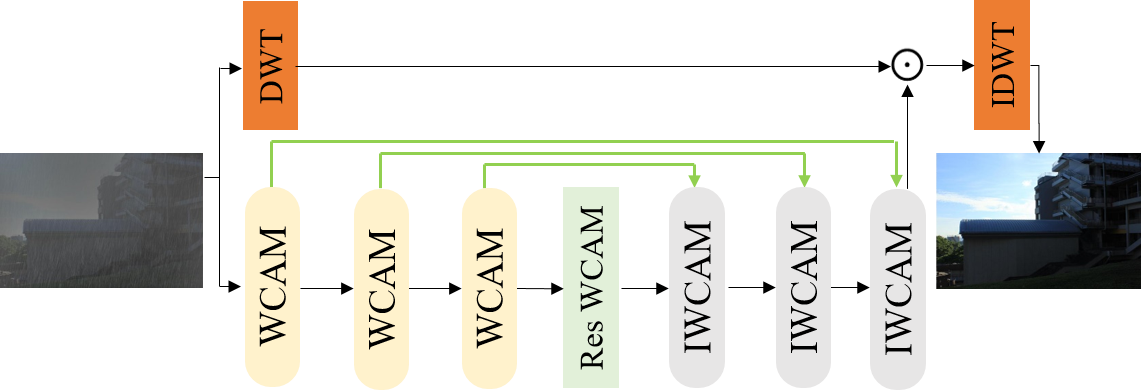}
\end{minipage}%
}%

\subfigure[]{
\begin{minipage}[t]{0.5\linewidth}
\centering
\includegraphics[width=1.5in]{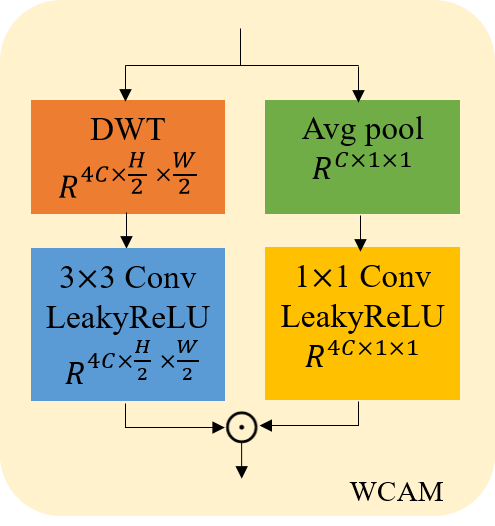}
\end{minipage}%
}%
\subfigure[]{
\begin{minipage}[t]{0.5\linewidth}
\centering
\includegraphics[width=1.5in]{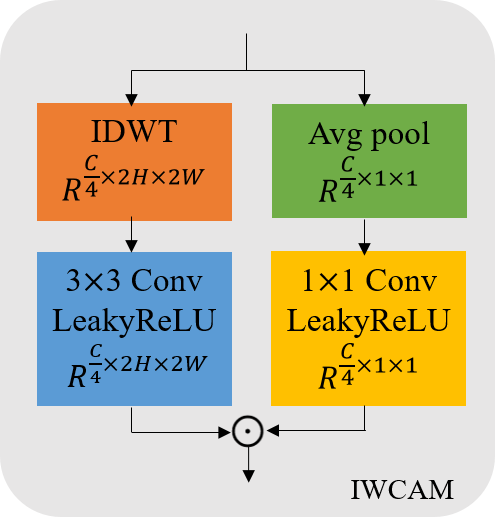}
\end{minipage}%
}%

\subfigure[]{
\begin{minipage}[t]{0.8\linewidth}
\centering
\includegraphics[width=3in]{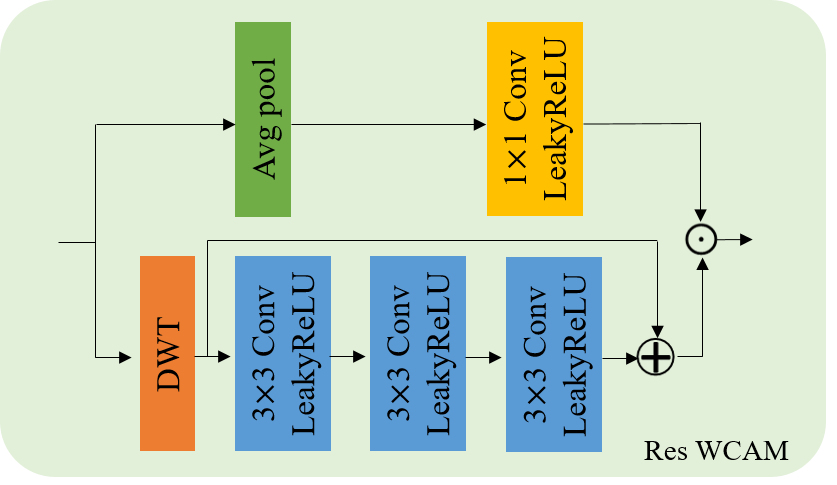}
\end{minipage}%
}%

\label{fig:figure20}
\centering
\caption{The proposed WCAM with a fusion network and components in the network. (a) The entire network, where green arrows mean skipping connections. (b) WCAM. (c) IWCAM. (d) The residul WCAM.}
\end{figure}

\subsection{Network Architecture}
Our network is encoder-decoder structure. The encoder consists of three WCAMs. Once a WCAM is adopted, the sizes of feature maps become quarter and the number of channels becomes four times, which not only captures multi-scale features but various frequency information. At the bottom of the network, the residual block \cite{he2016deep} combining wavelet channel attention is used and shown in Fig. 1(d). This module aggregates features and makes the learning process effective, especially in the event of deeper networks. Our decoder consists of three IWCAMs to generate clear images from extracted features. The terminal outputs are four confidence maps instead of the restored images. Once confidence maps for the derived wavelet inputs are predicted, they are multiplied by the four derived inputs to give the final derained image:
\begin{equation}
J= {\rm IDWT}(C_{LL} \odot I_{LL},C_{HL} \odot I_{HL},C_{LH} \odot I_{LH},C_{HH} \odot I_{HH})
\label{eq:res}
\end{equation}
where $C_{LL}$, $C_{HL}$, $C_{LH}$ ,and $C_{HH}$ are confidence maps for $I_{LL},I_{HL},I_{LH}$ and $I_{HH}$, respectively.
The reason for using the fusion mechanism is that the low-frequency sub-band plays a role in the objective quality, while the high-frequency sub-bands can affect the perceptual quality significantly \cite{Deng_2019_ICCV}. When low-frequency parts and high-frequency parts are optimized separately, the increase of objective quality cannot decrease perceptual quality. Additionally, like U-net \cite{ronneberger2015u}, we apply skipping connections to combine the identical size feature maps from WCAMs and IWCAMs so that the learning process converges quickly. The entire network architecture is shown in Fig. 1(a).
\vspace{-5mm}
\section{EXPERIMENTAL RESULTS}

\begin{figure*}[t]
  \centering
\includegraphics[width=\linewidth]{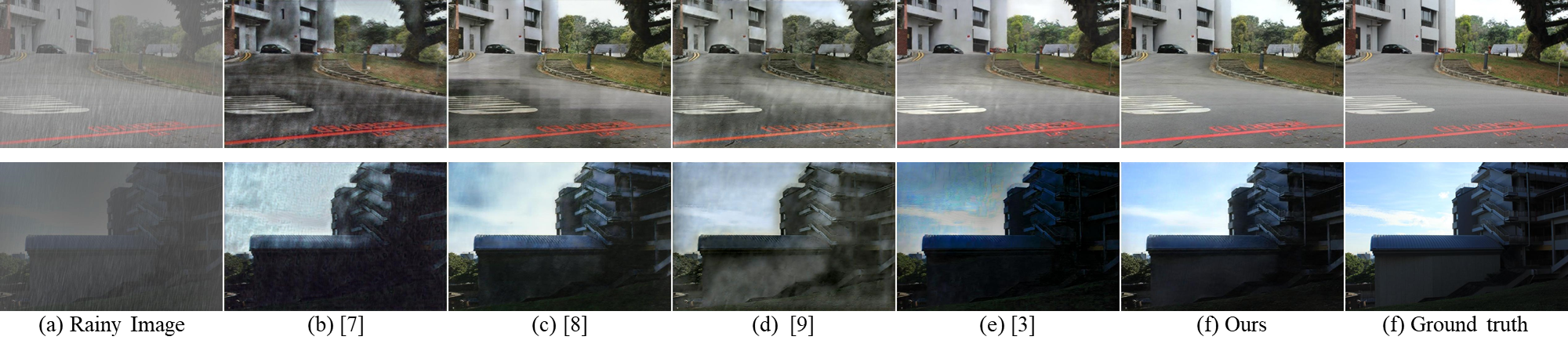}
\vspace{-0.8cm}
    \caption{{Derained results on sample images from the synthetic dataset \emph{Outdoor-Rain}. (Please zoom-in at screen to view details)
}}
    \vspace{-0.2cm}
\label{fig:figure2}
\end{figure*}

\begin{figure*}[t]
  \centering
\includegraphics[width=\linewidth]{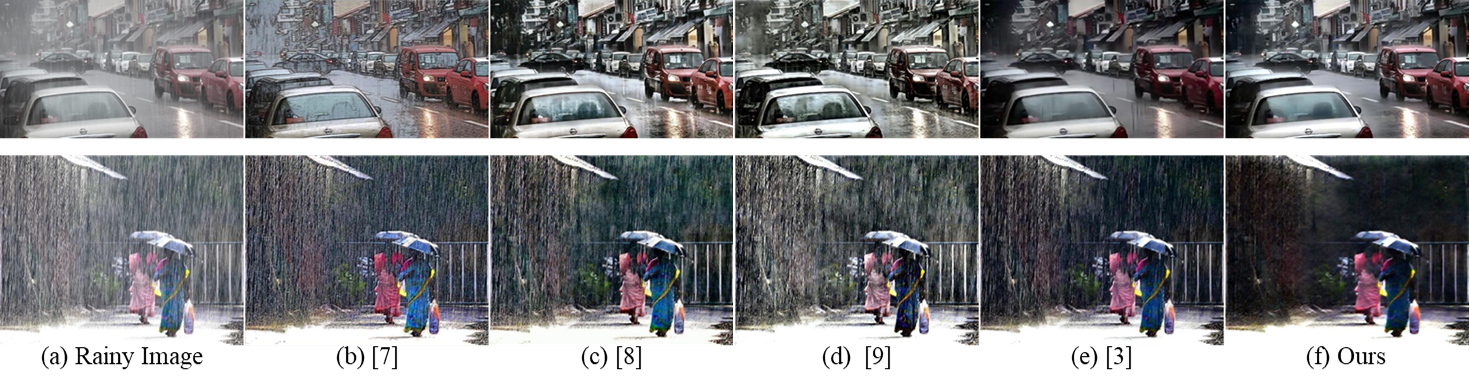}
\vspace{-0.8cm}
    \caption{{Derained results on real rainy images. (Please zoom-in at screen to view details)}}
\label{fig:figure2}
\vspace{-0.5cm}
\end{figure*}

\subsection{Datasets and training details}
In this work, \emph{Outdoor-Rain} dataset \cite{li2019heavy} is adopted to train and test the network. This dataset contains clear images, and the corresponding rainy images generated by Eq.(1). There are 7500 training and validation samples and 1,500 samples for evaluation. Both clean and rainy images are $480 \times 720$.
During training, images are cropped to $256 \times 256$, the wavelet SSIM loss \cite{yang2020net} and the L1 loss are employed, and RAdam \cite{liu2019radam} is used as an optimization algorithm with a mini-batch size of 16. The learning rate starts from 0.0001 and is divided by ten after 100 epochs. The models are trained for 300 iterations. The entire experiments are performed by the Pytorch framework.

\subsection{Image Deraining results}
PSNR and SSIM are chosen as objective metrics for quantitative evaluation. We select four state-of-the-art works \cite{pan2018learning, Zhang_2018_CVPR, fu2019lightweight,li2019heavy}  as deep learning-based benchmarks to make fairly comparisons with our purposed method. For the fair comparison, all methods are retrained on the same dataset. The comparison results are shown in Table 1. Table 1 presents our method has the largest PSNR and SSIM values among all deraining networks, which demonstrates our method has a superior performance of restoring clean images for this dataset and the frequency features are beneficial for restoring rainy images.
Furthermore, we perform various methods on synthetic and real rainy photos and results are depicted in Fig. 2 and Fig. 3. As revealed in Fig. 2 and Fig. 3, since comparative methods tend to miscalculate rainy concentration, restored images are dark or have remaining rain and mist. In contrast, our purposed method predicts better-derained results with balanced colors and detailed edges.

We analyze how the WCAM and the fusion help to refine derained results with three experiments.
The first experiment uses convolutions and wavelets without channel attention and fusion. The second experiment uses the proposed modules without fusion. The third experiment estimates feature maps to fuse sub-band images without channel attention. Table 2 compares our method against three baselines and demonstrates wavelet channel attention and fusion contribute the best results.
\vspace{-0.3cm}
\begin{table}[h]

  \centering
    \caption{{Quantitative SSIM and PSNR on the synthetic \emph{Outdoor-Rain} dataset.}}

\begin{tabular}{|lccccc|}
\hline
     & {\cite{pan2018learning}} & {\cite{Zhang_2018_CVPR}} & {\cite{fu2019lightweight}} & {\cite{li2019heavy}} & Ours  \\ \hline
PSNR & 17.92   & 21.64   & 18.16    & 21.17   & \textbf{24.89} \\ \hline
SSIM & 0.676   & 0.788   & 0.723    & 0.742   & \textbf{0.813} \\ \hline
\end{tabular}
\end{table}
\vspace{-7mm}
\begin{table}[ht!]

  \centering
    \caption{{Ablation study shows how the WCAM and the fusion help to refine derained results.}}

\begin{tabular}{|c|c|c|c|}
\hline
              & PSNR & SSIM  \\ \hline
Ours, w/o attention, w/o fusion & 21.44   & 0.784    \\ \hline
Ours, w/o fusion      & 23.34   &  0.764  \\ \hline
Ours, w/o attention      & 23.24   &  0.810 \\ \hline
\textbf{Ours} & \textbf{24.89}     & \textbf{0.813}        \\ \hline
\end{tabular}
\end{table}
\vspace{-0mm}
\section{CONCLUSION}
In this paper, the wavelet channel attention module with a fusion network is proposed for single image deraining. The wavelet transform and the inverse wavelet transform are substituted for down-sampling and up-sampling to extract various frequency features. The channel attention effectively controls ratios of feature maps. Furthermore, the proposed network estimates confidence maps for each derived wavelet input. Confidence maps and derived inputs are fused to render final derained results. Experiments on synthetic and real images verify the superiority of our model compared to the state-of-the-art results.
\bibliographystyle{IEEEbib}
\bibliography{icme2020template}

\begin{thebibliography}{10}

\bibitem{Li_2019_CVPR}
Siyuan Li, Iago~Breno Araujo, Wenqi Ren, Zhangyang Wang, Eric~K. Tokuda,
  Roberto~Hirata Junior, Roberto Cesar-Junior, Jiawan Zhang, Xiaojie Guo, and
  Xiaochun Cao,
\newblock ``Single image deraining: A comprehensive benchmark analysis,''
\newblock in {\em The IEEE Conference on Computer Vision and Pattern
  Recognition (CVPR)}, June 2019.

\bibitem{mccartney1976optics}
Earl~J McCartney,
\newblock ``Optics of the atmosphere: scattering by molecules and particles,''
\newblock {\em New York, John Wiley and Sons, Inc., 1976. 421 p.}, 1976.

\bibitem{li2019heavy}
Ruoteng Li, Loong-Fah Cheong, and Robby~T Tan,
\newblock ``Heavy rain image restoration: Integrating physics model and
  conditional adversarial learning,''
\newblock in {\em Proceedings of the IEEE Conference on Computer Vision and
  Pattern Recognition}, 2019, pp. 1633--1642.

\bibitem{chen2013generalized}
Yi-Lei Chen and Chiou-Ting Hsu,
\newblock ``A generalized low-rank appearance model for spatio-temporally
  correlated rain streaks,''
\newblock in {\em Proceedings of the IEEE International Conference on Computer
  Vision}, 2013, pp. 1968--1975.

\bibitem{luo2015removing}
Yu~Luo, Yong Xu, and Hui Ji,
\newblock ``Removing rain from a single image via discriminative sparse
  coding,''
\newblock in {\em Proceedings of the IEEE International Conference on Computer
  Vision}, 2015, pp. 3397--3405.

\bibitem{ronneberger2015u}
Olaf Ronneberger, Philipp Fischer, and Thomas Brox,
\newblock ``U-net: Convolutional networks for biomedical image segmentation,''
\newblock in {\em International Conference on Medical image computing and
  computer-assisted intervention}. Springer, 2015, pp. 234--241.

\bibitem{pan2018learning}
Jinshan Pan, Sifei Liu, Deqing Sun, Jiawei Zhang, Yang Liu, Jimmy Ren, Zechao
  Li, Jinhui Tang, Huchuan Lu, Yu-Wing Tai, et~al.,
\newblock ``Learning dual convolutional neural networks for low-level vision,''
\newblock in {\em Proceedings of the IEEE conference on computer vision and
  pattern recognition}, 2018, pp. 3070--3079.

\bibitem{Zhang_2018_CVPR}
He~Zhang and Vishal~M. Patel,
\newblock ``Density-aware single image de-raining using a multi-stream dense
  network,''
\newblock in {\em The IEEE Conference on Computer Vision and Pattern
  Recognition (CVPR)}, June 2018.

\bibitem{fu2019lightweight}
Xueyang Fu, Borong Liang, Yue Huang, Xinghao Ding, and John Paisley,
\newblock ``Lightweight pyramid networks for image deraining,''
\newblock {\em IEEE transactions on neural networks and learning systems},
  2019.

\bibitem{yang2019wavelet}
Hao-Hsiang Yang and Yanwei Fu,
\newblock ``Wavelet u-net and the chromatic adaptation transform for single
  image dehazing,''
\newblock in {\em IEEE International Conference on Image Processing (ICIP)}.
  IEEE, 2019, pp. 2736--2740.

\bibitem{hu2018squeeze}
Jie Hu, Li~Shen, and Gang Sun,
\newblock ``Squeeze-and-excitation networks,''
\newblock in {\em Proceedings of the IEEE conference on computer vision and
  pattern recognition}, 2018, pp. 7132--7141.

\bibitem{Ren_2018_CVPR}
Wenqi Ren, Lin Ma, Jiawei Zhang, Jinshan Pan, Xiaochun Cao, Wei Liu, and
  Ming-Hsuan Yang,
\newblock ``Gated fusion network for single image dehazing,''
\newblock in {\em The IEEE Conference on Computer Vision and Pattern
  Recognition (CVPR)}, June 2018.

\bibitem{li2016rain}
Yu~Li, Robby~T Tan, Xiaojie Guo, et~al.,
\newblock ``Rain streak removal using layer priors,''
\newblock in {\em Proceedings of the IEEE conference on computer vision and
  pattern recognition}, 2016, pp. 2736--2744.

\bibitem{woo2018cbam}
Sanghyun Woo, Jongchan Park, Joon-Young Lee, and In~So~Kweon,
\newblock ``Cbam: Convolutional block attention module,''
\newblock in {\em Proceedings of the European Conference on Computer Vision
  (ECCV)}, 2018, pp. 3--19.

\bibitem{yang2020characterizing}
Chao-Han Yang, Jun Qi, Pin-Yu Chen, et~al.,
\newblock ``Characterizing speech adversarial examples using self-attention
  u-net enhancement,''
\newblock in {\em IEEE International Conference on Acoustics, Speech and Signal
  Processing (ICASSP)}. IEEE, 2020, pp. 3107--3111.

\bibitem{mnih2014recurrent}
Volodymyr Mnih, Nicolas Heess, et~al.,
\newblock ``Recurrent models of visual attention,''
\newblock in {\em Advances in neural information processing systems}, 2014, pp.
  2204--2212.

\bibitem{mallat1999wavelet}
St{\'e}phane Mallat,
\newblock {\em A wavelet tour of signal processing},
\newblock Elsevier, 1999.

\bibitem{he2016deep}
Kaiming He, Xiangyu Zhang, Shaoqing Ren, and Jian Sun,
\newblock ``Deep residual learning for image recognition,''
\newblock in {\em Proceedings of the IEEE conference on computer vision and
  pattern recognition}, 2016, pp. 770--778.

\bibitem{Deng_2019_ICCV}
Xin Deng, Ren Yang, Mai Xu, and Pier~Luigi Dragotti,
\newblock ``Wavelet domain style transfer for an effective
  perception-distortion tradeoff in single image super-resolution,''
\newblock in {\em The IEEE International Conference on Computer Vision (ICCV)},
  October 2019.

\bibitem{yang2020net}
Hao-Hsiang Yang, Chao-Han~Huck Yang, and Yi-Chang~James Tsai,
\newblock ``Y-net: Multi-scale feature aggregation network with wavelet
  structure similarity loss function for single image dehazing,''
\newblock in {\em IEEE International Conference on Acoustics, Speech and Signal
  Processing (ICASSP)}. IEEE, 2020, pp. 2628--2632.

\bibitem{liu2019radam}
Liyuan Liu, Haoming Jiang, Pengcheng He, Weizhu Chen, Xiaodong Liu, Jianfeng
  Gao, and Jiawei Han,
\newblock ``On the variance of the adaptive learning rate and beyond,''
\newblock in {\em Proceedings of the Eighth International Conference on
  Learning Representations (ICLR 2020)}, April 2020.

\end{thebibliography}

\end{document}